\pdfoutput=1
\documentclass[lettersize,journal,twoside]{IEEEtran}
\usepackage{amsmath,amsfonts}
\usepackage{algorithmic}
\usepackage{algorithm}
\usepackage{array}
\usepackage{textcomp}
\usepackage{stfloats}
\usepackage{url}
\usepackage{verbatim}
\usepackage{graphicx}
\usepackage{cite}
\usepackage{placeins}
\usepackage{stfloats}
\usepackage{caption}
\usepackage{subcaption}
\usepackage{pifont}
\usepackage{makecell}   
\usepackage{array}      
\usepackage[
    colorlinks=true,
    linkcolor=blue,
    citecolor=blue,
    urlcolor=blue
]{hyperref}
\usepackage{orcidlink}

\newcommand{\cmark}{\ding{51}}  
\newcommand{\xmark}{\ding{55}}  

\makeatletter
\renewcommand{\@cite}[2]{\textcolor{blue}{[{#1}\if@tempswa , #2\fi]}}
\makeatother

\newcommand{\eqbracketref}[1]{\textcolor{blue}{(}\ref{#1}\textcolor{blue}{)}}

\begin{document}

\title{Setup-Invariant Augmented Reality for Teaching by Demonstration with Surgical Robots}

\author{Alexandre Banks\textsuperscript{\orcidlink{0009-0000-9617-4715}}, Richard Cook \textsuperscript{\orcidlink{0000-0002-3399-5812}}, and
Septimiu E. Salcudean\textsuperscript{\orcidlink{0000-0001-8826-8025}}, \textit{Life Fellow, IEEE}
\thanks{This work was supported by the Natural Sciences and Research Council of Canada Graduate Scholarships -- Master's Program and the C.A. Laszlo Biomedical Engineering Chair held by Professor Salcudean. \textit{(Corresponding author: Alexandre Banks; e-mail: abanksga@unb.ca)}}

\thanks{All authors are with the University of British Columbia (UBC), Vancouver, BC V6T 1Z4, Canada. Alexandre Banks and Septimiu Salcudean are with the UBC Electrical and Computer Engineering Department; Richard Cook is with the UBC Department of Surgery. Ethics approval was received from the UBC Behavioural Research Ethics Board (\#H24-01947)}

\thanks{Supplementary material is available at: \href{https://github.com/AlexandreBanks6/dVSTEAR_Supplemental_Files.git}{dVSTEAR\_Supplemental\_Files.git}}
\thanks{Open-source implementation on the dVRK: \href{https://github.com/AlexandreBanks6/dV-STEAR_Public.git}{dV-STEAR\_Public.git}}
}

\markboth{Banks \MakeLowercase{\textit{et al.}}: Setup-Invariant Augmented Reality for Teaching by Demonstration with Surgical Robots}%
{Banks \MakeLowercase{\textit{et al.}}: Setup-Invariant Augmented Reality for Teaching by Demonstration with Surgical Robots}



\maketitle
\thispagestyle{empty} 

\begin{abstract}
Augmented reality (AR) is an effective tool in robotic surgery education as it combines exploratory learning with three-dimensional guidance. However, existing AR systems require expert supervision and do not account for differences in the mentor and mentee robot configurations. To enable novices to train outside the operating room while receiving expert-informed guidance, we present dV-STEAR: an open-source system that plays back task-aligned expert demonstrations without assuming identical setup joint positions between expert and novice. Pose estimation was rigorously quantified, showing a registration error of 3.86$\boldsymbol{\pm}$2.01mm. In a user study (N=24), dV-STEAR significantly improved novice performance on tasks from the Fundamentals of Laparoscopic Surgery. In a single-handed ring-over-wire task, dV-STEAR increased completion speed (p=0.03) and reduced collision time (p=0.01) compared to dry-lab training alone. During a pick-and-place task, it improved success rates (p=0.004). Across both tasks, participants using dV-STEAR exhibited significantly more balanced hand use and reported lower frustration levels. This work presents a novel educational tool implemented on the da Vinci Research Kit, demonstrates its effectiveness in teaching novices, and builds the foundation for further AR integration into robot-assisted surgery.

\end{abstract}


\section{Introduction}

\IEEEPARstart{T}{raining} programs and educational technologies have not kept pace with the growing demand for robot-assisted surgery (RAS) \cite{shaw2022current}. One study found that 60\% of surgical residents in the U.S. intending to conduct RAS report a lack of preparation \cite{wang_robotic_2021}. With over 6500 clinical systems worldwide \cite{rivero2023robotic}, it is critical that trainees have access to robot-specific preparation early in their training. To fully achieve the benefits of RAS, such as shorter hospitalization times, it must be tightly integrated into surgical training through improved educational tools and increased robot-specific mentorship \cite{casas-yrurzum_new_2023,hong_simulation-based_2021,qian_review_2020}.

The four primary methods for RAS training are: 1)~box-trainers, 2)~apprenticeship models, 3)~virtual reality (VR) simulators, and~4) augmented reality (AR) guidance \cite{lahanas_novel_2015}. In the first approach, box-trainers, novice surgeons develop basic skills through hands-on practice in a dry-lab setting with standardized training tasks \cite{hernansanz_endotrainer_2023,smith_fundamentals_2014,casas-yrurzum_new_2023}. Although box-trainers enable novices to learn through exploring, they have limitations (Table \ref{table:comparison_of_training}), including the lack of expert mentorship, instructions limited to written guidelines, and novices having to rely on prior experience \cite{wang_robotic_2021,hong_simulation-based_2021}.

\begin{table}[h!]
    \centering
    \caption{Comparison of RAS Training Methods.}
    \setlength{\tabcolsep}{0pt} 
    \begin{tabular}{>{\raggedright}p{1.75cm} >
    {\centering\arraybackslash}p{1.5cm} > {\centering\arraybackslash}p{1.5cm}>
    {\centering\arraybackslash}p{1.25cm}>
    {\centering\arraybackslash}p{1.85cm}>{\centering\arraybackslash}p{1.025cm}}
    \hline\hline
         & \makecell{\textbf{Real-World}\\\textbf{Exploration}} 
         & \makecell{\textbf{Expert}\\\textbf{Mentorship}} 
         & \makecell{\textbf{Pose}\\ \textbf{Guidance}} 
         & \makecell{\textbf{Practice}\\\textbf{Independently}} 
         & \makecell{\textbf{Tailors}\\ \textbf{Content}} \\
     \hline
     \noalign{\vskip 2pt}
        Box-Trainers & \cmark & \xmark & \xmark & \cmark & \cmark \\ 
       Apprenticeship & \cmark & \cmark & \xmark & \xmark & \xmark \\ 
        VR Simulators & \xmark & \xmark & \cmark & \cmark & \cmark \\ 
        Traditional AR & \cmark & \cmark & \cmark & \xmark & \cmark \\ 
        \textbf{dV-STEAR} & \cmark & \cmark & \cmark & \cmark & \cmark \\ 
        \hline
    \end{tabular}
    \label{table:comparison_of_training}
\end{table}


Novices often transition from box-trainers directly to an apprenticeship model where they gain practical skills intraoperatively \cite{jarc_proctors_2017,wang_robotic_2021,casas-yrurzum_new_2023}. During apprenticeship, 
instruction is either provided as verbal guidance \cite{lahanas_novel_2015} or through a dual-console system where the expert can control the arms and guide the novice \cite{wang_robotic_2021}. While this approach provides direct mentorship, it is essential that the novice is competent prior to controlling the robot and can practice outside the OR \cite{abboudi_current_2013}. Another limitation is that the apprenticeship model is typically constrained to verbal guidance  \cite{lahanas_novel_2015} and lacks customization to an individual's learning needs. In addition, surgeons often face burnout \cite{galaiya2020factors}, making intraoperative mentorship an added burden on physicians \cite{rutter2002stress}. As a result, there is a need for training methods where novice surgeons can practice outside the OR and still benefit from expert-level guidance.

To overcome limitations in box-training and apprenticeship, VR simulators were developed so novices could receive tailored training content, practice at any time, and receive textual or virtual guidance \cite{hong_simulation-based_2021,wang_robotic_2021}. Many commercial VR platforms exist, including 
SimNow (Intuitive Surgical, Inc), which is native to da Vinci robots. VR simulation, however, lacks expert mentorship and real-world exploration, making box-trainers often better for advanced skill development \cite{raison2020virtually}.

Augmented reality has emerged as a promising approach for RAS education \cite{fu2023recent,ritter_concurrent_2007} as it addresses limitations in VR by immersing the user in hands-on training. Additionally, AR offers a risk-free environment, virtual guidance, and a learner-centered approach \cite{qian_review_2020,hong_simulation-based_2021}. By overlaying guidance information, AR has the potential to enhance surgical education beyond traditional approaches and provide novice surgeons with an engaging and flexible learning environment.

The most widespread application of AR within RAS is intraoperative guidance and surgical planning \cite{seetohul_augmented_2023,fu2023recent}. For example, many systems superimpose preoperative scans onto the surgical scene, allowing the surgeon to easily view medical images \cite{volonte_augmented_2011,seetohul_augmented_2023}. 
Other AR frameworks guide trocar placement in RAS \cite{falk_cardio_2005}, delineate tumor margins for resection \cite{singla_intra-operative_2017}, and superimpose visual force feedback \cite{durat2023enhancing}. Head-mounted displays have also been introduced to support surgical assistants in arm placement and suctioning \cite{qian_arssist_2018}. Although AR is used to facilitate OR workflows and overlay multi-modal information on the surgical scene, its application in training remains limited. A recent survey article reported that only 2 of 49 papers on AR in RAS were for training purposes \cite{qian_review_2020}. 

Early work in AR-enabled RAS training (AR-RAST) showed that experts performed significantly better than novices on an AR peg transfer task and that both groups found AR to be realistic \cite{lahanas_novel_2015}. This study, however, did not use AR to explicitly guide trainees. Other systems enable a remote expert to guide a virtual pointer through a head-mounted display \cite{long_integrating_2022} or 2D laparoscopic screen \cite{feng_virtual_2018}. Although these 2D telementoring systems improve the novice's economy of movement \cite{feng_virtual_2018}, they require an expert to be present. They are also setup-dependent, meaning they do not account for changes in the robot's setup joints --- the first six un-actuated joints used to manually position each arm. Moreover, these systems do not provide information on the 6 degree of freedom (6 DoF) instrument pose, which is the preferred method of guidance among both mentors and trainees \cite{jarc_beyond_2016}.

To provide information on tool orientation, some AR-RAST systems overlay 3D semi-transparent ghost instruments onto a stereoscopic view \cite{jarc_proctors_2017,jarc_beyond_2016,matu_stereoscopic_2014,shabir_towards_2021,shabir_preliminary_2022}. Novices can follow these 3D renderings with 3mm accuracy \cite{shabir_preliminary_2022} and ghost instruments uniquely facilitate the guidance of finger pinch, wrist rotation, and bimanual hand motions \cite{jarc_proctors_2017}. However, these telementoring systems still require an expert to be in the loop, limiting opportunities for novices to train independently. Additionally, because the instrument's pose is not located with respect to the scene, the novice's and mentor's robot must be placed in a similar configuration with the expert compensating for offsets in 3D tool positioning. These direct guidance systems also lack quantification of their 6 DoF pose estimation accuracy and have not been evaluated for their impact on novice learning. 

More recently, artificial intelligence (AI) has been merged with AR to generate guidance for novices' training on dry-lab tasks \cite{long_robotic_2023}. This system removes dependence on an expert for mentorship but requires that objects be placed in the same location with respect to the endoscope and only provides 2D trajectory guidance. Another approach to enable novices to practice independently involves recording surgeon motions as hand gestures \cite{cota_neto_what_2024} 
or as videos of the surgical scene \cite{krauthausen_robotic_2021} and then overlaying them at a later time point. While promising, these methods do not render articulated 3D ghost instruments and are setup-dependent, requiring that the robot's setup joints remain unchanged between recording and playback.

Most existing AR-RAST systems are constrained by the need for an expert to be present. The few technologies eliminating direct mentorship face limitations, including the inability to register motions to a structured surgical training task and the requirement that the robot setup joints remain the same for the expert and novice. To address these issues, we present dV-STEAR: an AR application that records expert motions relative tasks from the Fundamentals of Laparoscopic Surgery \cite{smith_fundamentals_2014} and replays task-aligned guidance to the novice in any robot configuration. 

The main contributions of this work are: 1) an AR framework for markerless setup-invariant, playback of expert demonstrations; 2) a novel method for back-propagating virtual objects to render articulated surgical instruments; 3) systematic quantification of pose estimation and AR rendering errors; 4) a user study (N=24) evaluating the impact of dV-STEAR on novice performance; and 5) an open-source implementation on the da Vinci Research Kit (dVRK) for ease of modification, available at: \href{https://github.com/AlexandreBanks6/dV-STEAR_Public.git}{dV-STEAR\_Public.git}.

Methods for dV-STEAR are detailed in Section \ref{sec:methods} with Section \ref{subsec:pose_estim} describing the markerless patient-side manipulator (PSM) pose estimation. Scene registration is presented in Section \ref{subsec:scene_reg}, followed by AR playback in Section \ref{subsec:rendering_andplayback}. Methods for kinematic error compensation and hand-eye calibration are in Sections \ref{subsec:error_corr} and \ref{subsec:cam_calib}. System errors are quantified in Section \ref{subsec:pose_estim}, and user study results of the impact of dV-STEAR on novice learning are in Section \ref{subsec:results_userstudy}. The paper concludes with a discussion of results and avenues for future work in Sections \ref{sec:Discussion} and \ref{sec:Conclusion}. 

\section{Methods}
\label{sec:methods}
\begin{figure}[h!]
    \begin{center}   
    \includegraphics[trim={0.4cm 19.8cm 3.05cm 0.8cm},clip,width=\columnwidth]{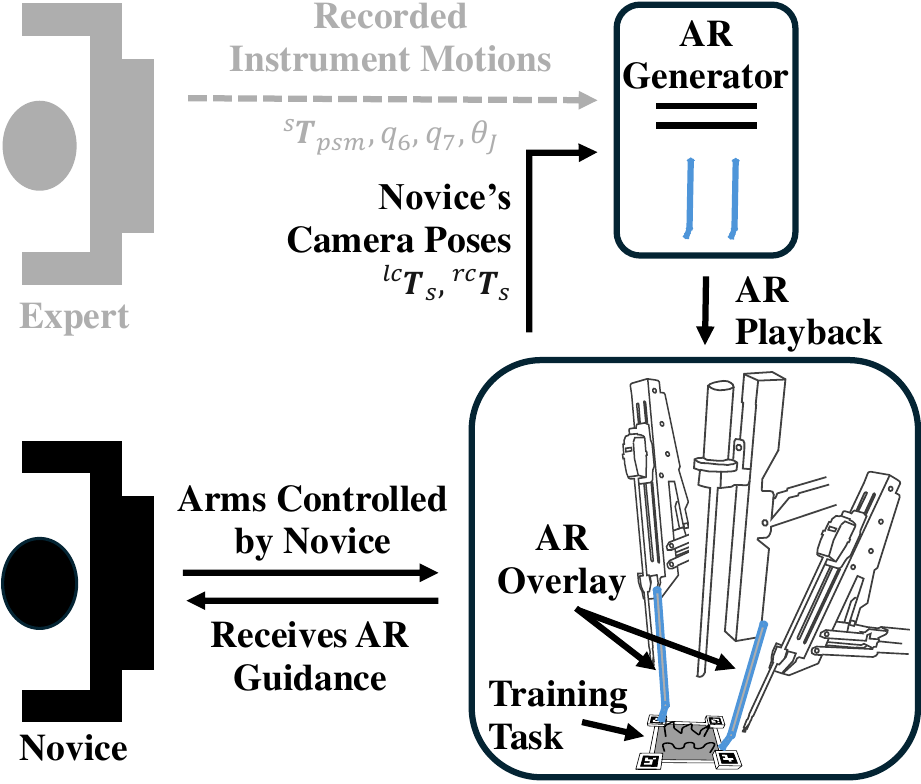}
    \caption{Overview of dV-STEAR for setup-invariant record and playback of AR tools in RAS training. Expert surgeon motions are recorded as PSM end-effector pose relative to the training task, ${}^{s}\boldsymbol{T}_{psm}$, and the instrument's final three joint parameters, $q_6,q_7,\theta_J$. These are played back to the novice at a later time. The view during AR rendering corresponds to the left/right ECM camera poses, ${}^{lc}\boldsymbol{T}_{s},{}^{rc}\boldsymbol{T}_{s}$, relative to the training task.}
    \label{fig:methods_overview_1}
    \end{center}
\end{figure}

Fig. \ref{fig:methods_overview_1} gives an overview of dV-STEAR. PSM poses and joint states are recorded as an expert completes a training task, and these are later played back to novices as a 3D overlay. During AR rendering, the endoscopic camera's (ECM) left and right camera poses serve as viewpoints for replaying PSM motions, decoupling the setup joints from AR playback. 

\begin{figure}[h!]
    \begin{center}   
    \includegraphics[trim={0.4cm 9.7cm 3.05cm 14.7cm},clip,width=\columnwidth]{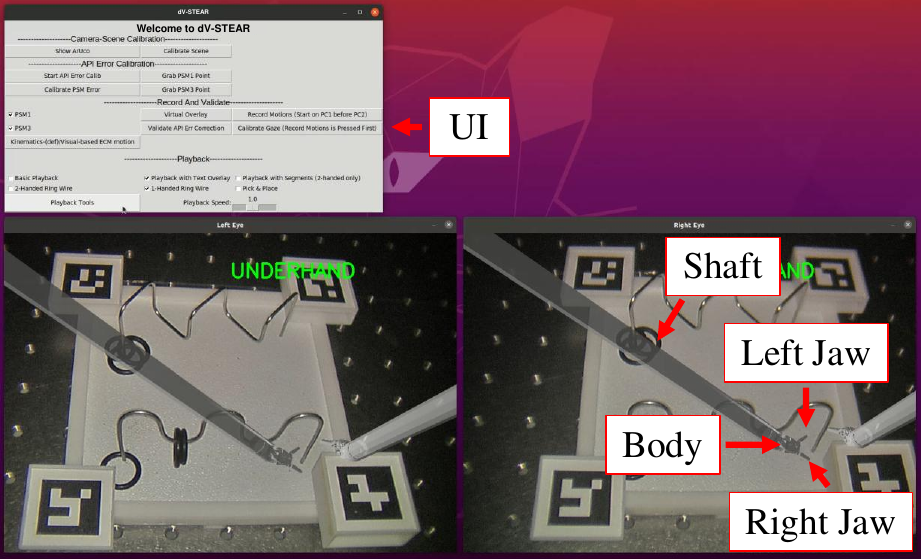}
    \caption{dV-STEAR UI, showing the four components of the articulated instruments: left jaw, right jaw, body, and shaft.}
    \label{fig:methods_overview_2}
    \end{center}
\end{figure}

dV-STEAR includes a user interface (UI) for controlling the AR renderings and playback speed (Fig. \ref{fig:methods_overview_2}), enabling users to tailor the pace of training. This real-time platform is implemented on the dVRK, interfaced via the Robotic Operating System (ROS), and AR renderings run on a Ubuntu 20.04 Linux computer. Our dVRK consists of two da Vinci S\textsuperscript{\textregistered} PSMs (PSM1 and PSM3) and a da Vinci Si\textsuperscript{\textregistered} ECM. Stereo endoscope frames are captured by the Linux PC using a Blackmagic Design\textsuperscript{\textregistered} DeckLink Duo 2 frame grabber card. Processed stereoscopic video with AR overlay is displayed on the dVRK's left/right monitors. AR rendering is performed with ModernGL and image processing is implemented with the Open Source Computer Vision (OpenCV) library. Matrix operations use the OpenGL Mathematics (GLM) library accelerated with a NVIDIA\textsuperscript{\textregistered} GTX 1660. 

For simplicity, equations refer to a single PSM, though they apply to both PSM1 and PSM3. Matrices are denoted by uppercase bold symbols, vectors by lowercase bold, and scalars by lowercase unbolded. ${}^{x}\boldsymbol{T}_{y}$ represents a homogeneous transformation mapping 3D points $\boldsymbol{p}^y$ in coordinate system $\{\underline{\boldsymbol{C}}_{y},\boldsymbol{o}_{y}\}$ to points $\boldsymbol{p}^x$ in coordinate system $\{\underline{\boldsymbol{C}}_{x},\boldsymbol{o}_{x}\}$.

\subsection{End-Effector Pose Estimation and Recording}\label{subsec:pose_estim}

\begin{figure}[h!]
\begin{center}    \includegraphics[width=\columnwidth,trim={9.5cm 2.2cm 8.3cm 1.2cm},clip]{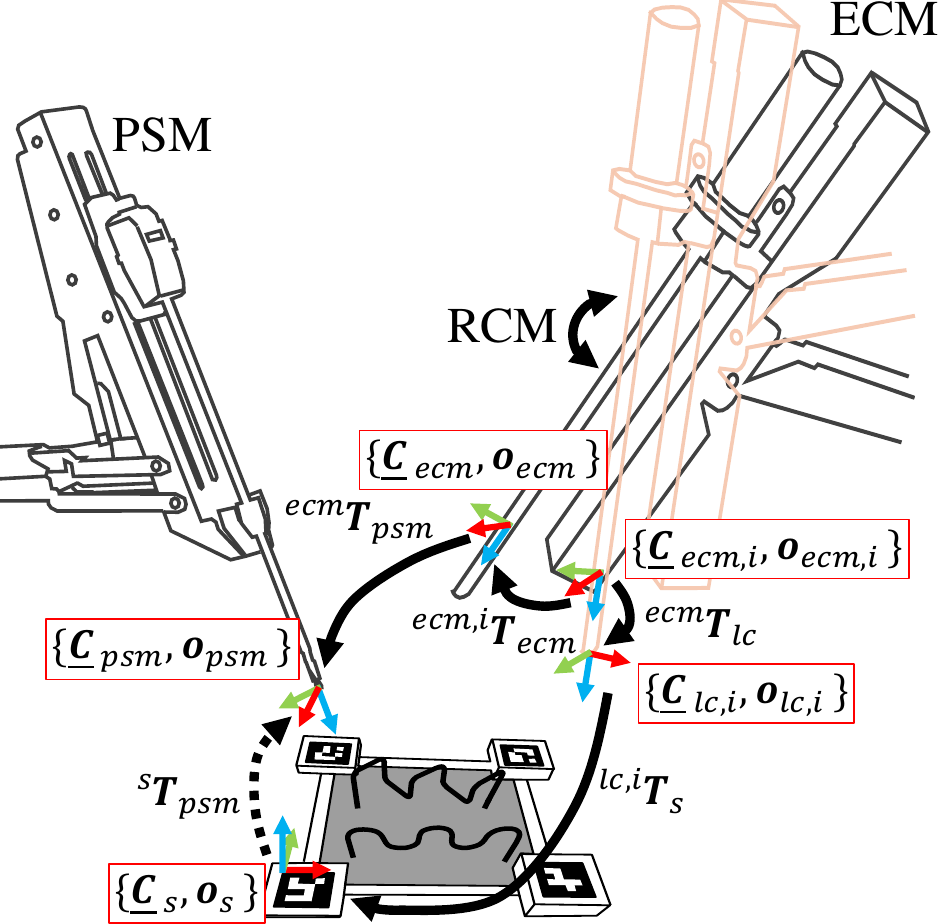}
    \caption{Overview of the transforms to estimate PSM pose during expert surgeon motion recording. ${}^{lc,i}\boldsymbol{T}_{s}$ is the registration from the initial camera pose to the dry-lab surgical scene. ${}^{ecm}\boldsymbol{T}_{lc}$ and ${}^{ecm,i}\boldsymbol{T}_{ecm}$ are the hand-eye and ECM motion transforms, respectively. ${}^{ecm}\boldsymbol{T}_{psm}$ is the uncorrected transform from the ECM to the PSM given by the dVRK API. ${}^{s}\boldsymbol{T}_{psm}$ is the resulting transform recorded when the surgeon completes tasks.}    \label{fig:recording_pose}
\end{center}
\end{figure}

To ensure the demonstrations from the expert surgeon are not a function of robot setup joints, the pose of the PSM, $\{\underline{\boldsymbol{C}}_{psm},\boldsymbol{o}_{psm}\}$, is recorded with respect to the surgical training task, $\{\underline{\boldsymbol{C}}_{s},\boldsymbol{o}_{s}\}$, as shown by the dotted line in Fig. \ref{fig:recording_pose}. The homogeneous transformation from the training task to the PSM,  ${}^{s}\boldsymbol{T}_{psm}$, is computed as a function of scene registration, ECM motion, hand-eye calibration, and API-reported PSM pose with error correction.

The first term in the PSM pose estimation equation \eqbracketref{eqn:PoseRecording} is the transform from the initial left camera pose to the scene, ${}^{lc,i}\boldsymbol{T}_{s}$. This is computed by registering the left camera to the scene at the beginning of the recording. Registration is done monocularly; therefore, the choice of using the left or right endoscopic camera is arbitrary. Section \ref{subsec:scene_reg} gives more details on this robust, monocular registration method.

After initial scene registration, the ECM is allowed to move freely and the transform from the initial ECM frame, $\{\underline{\boldsymbol{C}}_{ecm,i},\boldsymbol{o}_{ecm,i}\}$, to the current ECM frame, $\{\underline{\boldsymbol{C}}_{ecm},\boldsymbol{o}_{ecm}\}$, is computed as ${}^{ecm,i}\boldsymbol{T}_{ecm}={}^{base}\boldsymbol{T}_{ecm,i}^{-1} \; {}^{base}\boldsymbol{T}_{ecm}$. The matrix ${}^{base}\boldsymbol{T}_{ecm,i}$ is the transform from the robot base to the initial ECM pose during scene registration, and ${}^{base}\boldsymbol{T}_{ecm}$ is the current ECM pose. The transformation from ECM to the PSM, ${}^{ecm}\boldsymbol{T}_{psm}$, is found using the dVRK software interface. Because of cable stretch and shaft deflection, there are errors in the dVRK API reported PSM pose. A correction term is used to account for the API error, $\boldsymbol{T}_{cor}$, with further details given in section \ref{subsec:error_corr}. Finally, ${}^{ecm}\boldsymbol{T}_{lc}$ is the hand-eye transform computed through a dual quaternion approach described in \ref{subsec:cam_calib}.

\begin{equation}
\label{eqn:PoseRecording}
{}^{s}\boldsymbol{T}_{psm}={}^{lc,i}\boldsymbol{T}_{s}^{-1}\; \boldsymbol{T}_{cor}\; {}^{ecm}\boldsymbol{T}_{lc}^{-1}\; {}^{ecm,i}\boldsymbol{T}_{ecm}\; {}^{ecm}\boldsymbol{T}_{psm} 
\end{equation}

An implicit assumption in this PSM tracking strategy is that the task is a rigid body. Our contribution is not focused on registration methods for non-rigid objects, but rather on implementing this setup-invariant playback approach on a real robotic platform, robustly quantifying system errors, and evaluating this novel concept through a user study. Furthermore, while fiducials are used for the initial scene registration, our method does not require markers on the PSM for pose estimation.
This ensures a markerless approach.

\subsection{Surgical Training Task Registration} \label{subsec:scene_reg}
\begin{figure}[h!]
\begin{center}    \includegraphics[width=\columnwidth,trim={10.8cm 6cm 9.1cm 2.8cm},clip]{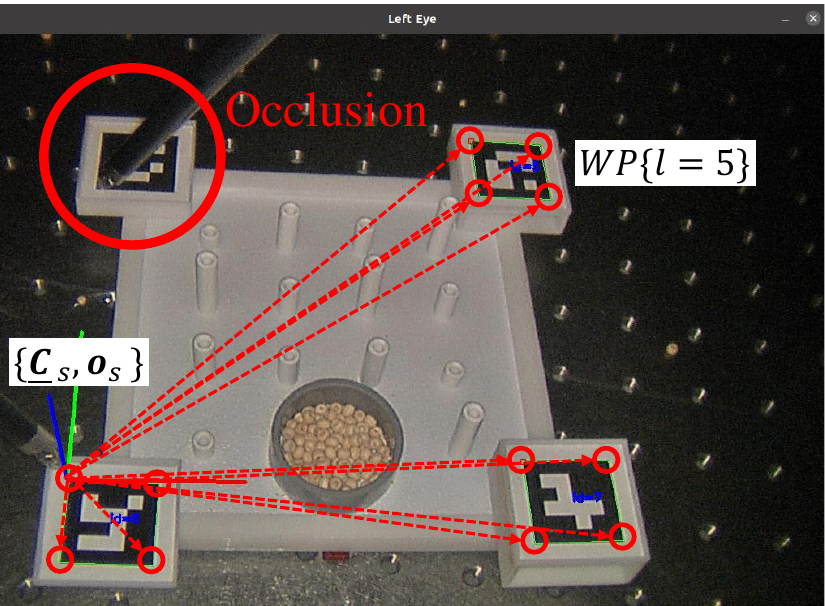}
    \caption{\textit{Pick-and-Place} dry-lab surgical task with mounts for ArUco markers. Depicted are visible 3D corner positions in the object's coordinate frame $\{\underline{\boldsymbol{C}}{s},\boldsymbol{o}_{s}\}$. The registration algorithm computes the ${}^{c,i}\boldsymbol{T}_{s}$ transform under occlusion so long as one ArUco is visible.}    \label{fig:scene_reg}
\end{center}
\end{figure}

We present a robust method for registering the endoscopic cameras to the dry-lab surgical task, Fig. \ref{fig:scene_reg}. This efficient registration is performed at the start of recording the expert surgeon's motions and before replaying them to a novice. 

Algorithm \ref{alg:scene_reg} estimates the camera-to-scene transformation, ${}^{c,i}\boldsymbol{T}_{s}$, for both the left and right cameras. The first step (lines 3-5) is to loop for $N_{frame,thres}$ frames, detecting candidate 2D corners, $\boldsymbol{C}$, and ArUco labels, $\boldsymbol{L}$, in the current frame, $\boldsymbol{I}$, using the implementation of \cite{garrido2014automatic} in OpenCV. Each ArUco marker has four corners and a unique label. Only 2D corners with labels matching the known ArUco $IDs$ are appended to an array of image plane points, $\boldsymbol{T}_{im,pts}$ (lines 6-8). The 3D location of these corners in the scene’s coordinate frame, $\{\underline{\boldsymbol{C}}{s},\boldsymbol{o}_{s}\}$, are retrieved from a lookup table, $WP\{\cdot\}$, and stored in array $\boldsymbol{T}_{s,pts}$, giving 2D-to-3D point correspondences (lines 11-12). If the number of ArUco detections exceeds $N_{detect,thres}$, the point correspondences are then passed to the MAGSAC perspective-n-point (PnP) solver, $f_{pnp}$, described in \cite{barath2019magsac}. This gives the final camera-to-scene transform ${}^{c,i}\boldsymbol{T}_{s}$. In our implementation, $N_{f,thres}=N_{d,thres}=10$.

\begin{algorithm}
    \caption{Occlusion Robust Scene Registration}\label{alg:scene_reg}
    \begin{algorithmic}[1]
        \STATE $n_{detect} \gets 0$
        \STATE $\boldsymbol{T}_{im,pts},\boldsymbol{T}_{s,pts} \gets \{\}$      
        
        \vspace{-0.6em}
        \rule{\linewidth}{0.5pt}  
        \vspace{-1.4em}        
        \FOR{$\boldsymbol{I} \;in \;N_{frame,thres}$}
            \STATE $\boldsymbol{C},\boldsymbol{L} \gets f_{detect}(I)$
            \FOR{$l \; in \; L \; in \; IDs$}
                \STATE $\boldsymbol{T}_{im,pts} \gets \text{append}(\boldsymbol{T}_{im,pts}, \boldsymbol{C}\{l\})$
                \STATE $\boldsymbol{T}_{s,pts} \gets \text{append}(\boldsymbol{T}_{s,pts},WP\{l\})$
                \STATE $n_{detect} \gets n_{detect}+1$
            \ENDFOR        
        \ENDFOR
        
        \vspace{-0.6em}
        \rule{\linewidth}{0.5pt}  
        \vspace{-1.4em}        
        \IF{$n_{detect}\geq N_{d,thres}$}
            \STATE ${}^{c,i}\boldsymbol{T}_{s} \gets f_{pnp}(\boldsymbol{T}_{s,pts},\boldsymbol{C}_{im,pts})$
        \ENDIF
    \end{algorithmic}
\end{algorithm}

\subsection{Tool Rendering and Playback} \label{subsec:rendering_andplayback}
Expert demonstrations are played back to the novice using the semi-transparent ghost tools shown in Fig. \ref{fig:methods_overview_2} These CAD models of the da Vinci S\textsuperscript{\textregistered} large needle drivers, consist of four rigid components: shaft, body, and left/right jaws. Each component is represented as a set of vertices, $\boldsymbol{p}^{loc}$, in the model's local coordinate system. During playback, these 3D model coordinates are projected into normalized 2D image coordinates, $\boldsymbol{p}^{rend}$, using the rendering pipeline in \eqbracketref{eqn:rendering_pipeline} below.

\begin{equation}
\label{eqn:rendering_pipeline}
\boldsymbol{p}^{rend}={}^{rend}\boldsymbol{T}_{view}\;{}^{view}\boldsymbol{T}_{s}\;{}^{s}\boldsymbol{T}_{loc}\;\boldsymbol{p}^{loc} 
\end{equation}

The transformation ${}^{s}\boldsymbol{T}_{loc}$ maps points in the component's local coordinate system to the surgical task frame, $\{\underline{\boldsymbol{C}}_{s},\boldsymbol{o}_{s}\}$. Since there are four CAD models for the ghost instruments, we define four unique model transformations in \eqbracketref{eqn:scene_to_local}. Each transform ${}^{s}\boldsymbol{T}_{x,loc}$ is a composition dependent on the pose and joint states of the PSM when recorded by the expert. Our instrument rendering approach is novel in that we use the pose of the end-effector, ${}^{s}\boldsymbol{T}_{psm}$, described in \eqbracketref{eqn:PoseRecording} and the recorded joint angles to move backward through the kinematic chain of the robot's wrist while placing local 3D bodies. This reduces propagation errors that arise from moving forward through the robot's first 5 degrees of freedom.

\begin{subequations}\label{eqn:scene_to_local}
\begin{align}
{}^{s}\boldsymbol{T}_{J_L,loc}&={}^{s}\boldsymbol{T}_{psm}\; {}^{psm}\boldsymbol{T}_{w}\;{}^{w}\boldsymbol{T}_{J_L,loc}\\
{}^{s}\boldsymbol{T}_{J_R,loc}&={}^{s}\boldsymbol{T}_{psm}\; {}^{psm}\boldsymbol{T}_{w}\;{}^{w}\boldsymbol{T}_{J_R,loc}\\
{}^{s}\boldsymbol{T}_{b,loc}&={}^{s}\boldsymbol{T}_{psm}\; {}^{psm}\boldsymbol{T}_{w}\;{}^{b,loc}\boldsymbol{T}_{w}^{-1}\\
{}^{s}\boldsymbol{T}_{s,loc}&={}^{s}\boldsymbol{T}_{psm}\; {}^{psm}\boldsymbol{T}_{w}\;{}^{b,loc}\boldsymbol{T}_{w}^{-1}\;{}^{sh,loc}\boldsymbol{T}_{b,loc}^{-1}
\end{align}
\end{subequations}

\begin{figure}[h!]
    \centering
    \includegraphics[width=\columnwidth,trim={0.225cm 10cm 0.15cm 0.5cm},clip]{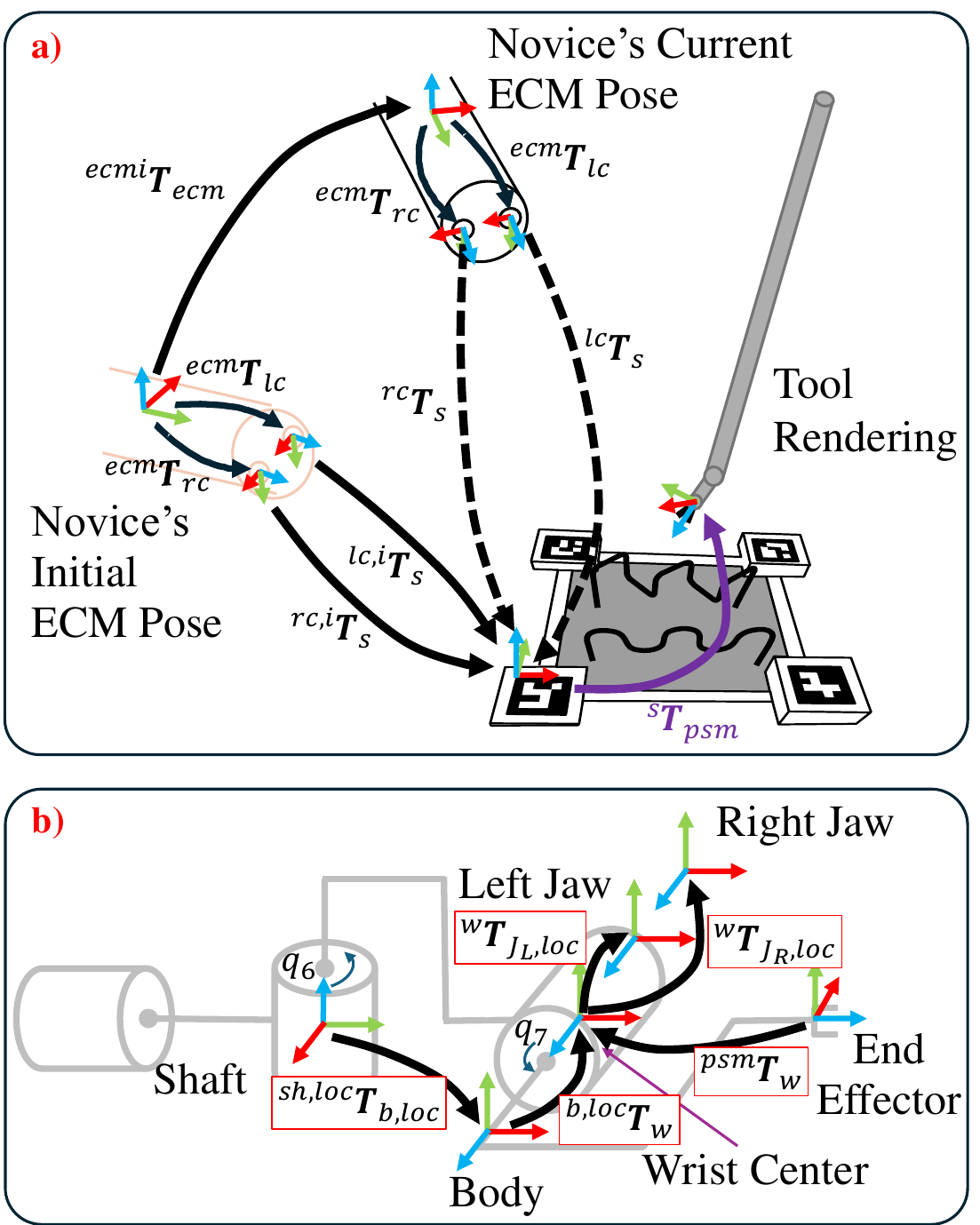}    
    \caption{\textbf{a)} Transforms during playback of the AR instruments. ${}^{lc,i}\boldsymbol{T}_{s}$ and ${}^{rc,i}\boldsymbol{T}_{s}$ are registrations from the initial left/right camera frames to the dry-lab surgical scene. ${}^{ecm}\boldsymbol{T}_{lc}$ and ${}^{ecm}\boldsymbol{T}_{rc}$ are the rigid hand-eye transforms. ECM motion is given by ${}^{ecm,i}\boldsymbol{T}_{ecm}$. The transform ${}^{s}\boldsymbol{T}_{psm}$ in purple is the pose of the PSM recorded from the expert surgeon and overlayed as an AR tool; \textbf{b)} Transformations used to move backwards through the PSM's kinematic chain and place the four rigid AR components: left jaw ($J_L,loc$), right jaw ($J_R,loc$), body ($b,loc$), and shaft ($sh,loc$).}
    \label{fig:playback}
\end{figure}

Here, ${}^{psm}\boldsymbol{T}_{w}$ is a translation from the PSM end-effector to the wrist center (shown in Fig. \ref{fig:playback}b.). The transformations ${}^{s}\boldsymbol{T}_{J_L,loc}$ and ${}^{s}\boldsymbol{T}_{J_R,loc}$ define the pose of the left and right jaws as a function of the jaw separation angle: $\pm \theta_J/2$. The body and shaft poses are defined by the instrument's Denavit-Hartenberg parameters such that ${}^{b,loc}\boldsymbol{T}_{w}={}^{b,loc}\boldsymbol{T}_{w}(q_7)$ and ${}^{sh,loc}\boldsymbol{T}_{b,loc}={}^{sh,loc}\boldsymbol{T}_{b,loc}(q_6)$.

To ensure recorded instrument motions are viewed from the novice's perspective, the view-to-scene transformation, ${}^{view}\boldsymbol{T}_{s}$, in \eqbracketref{eqn:rendering_pipeline} is dynamically aligned with the pose of the endoscopic cameras during playback. Since this is a stereo application, we define two view-to-scene transforms \eqbracketref{eqn:view_to_scene}. These camera frames, $\{\underline{\boldsymbol{C}}_{lc},\boldsymbol{o}_{lc}\}$ and $\{\underline{\boldsymbol{C}}_{rc},\boldsymbol{o}_{rc}\}$, are related to the surgical scene, $\{\underline{\boldsymbol{C}}_{s},\boldsymbol{o}_{s}\}$, by multiplying the hand-eye, scene registration, and ECM motion transforms as shown in Fig. \ref{fig:playback}a.

\begin{subequations}\label{eqn:view_to_scene}
\begin{align}
{}^{lc}\boldsymbol{T}_{s}&={}^{ecm}\boldsymbol{T}_{lc}^{-1}\; {}^{ecm,i}\boldsymbol{T}_{ecm}^{-1}\;{}^{ecm}\boldsymbol{T}_{lc}\;{}^{lc,i}\boldsymbol{T}_{s}\\
{}^{rc}\boldsymbol{T}_{s}&={}^{ecm}\boldsymbol{T}_{rc}^{-1}\; {}^{ecm,i}\boldsymbol{T}_{ecm}^{-1}\;{}^{ecm}\boldsymbol{T}_{rc}\;{}^{rc,i}\boldsymbol{T}_{s}
\end{align}
\end{subequations}

The final rendering step is to compute the projection matrix, ${}^{rend}\boldsymbol{T}_{view}$, which transforms 3D model vertices in the left/right camera coordinate frames into normalized 2D image coordinates. ${}^{rend}\boldsymbol{T}_{view}$ is derived from the standard pinhole camera model \cite{zhang2000flexible} and includes frustrum-based clipping to remove vertices outside the viewing volume. 


\subsection{Robot Kinematics Error Correction} \label{subsec:error_corr}
Because the patient-side manipulators of the dVRK have 13 joint parameters, the reported PSM pose, ${}^{ecm}\boldsymbol{T}_{psm}$, in equation \eqbracketref{eqn:PoseRecording} has large absolute errors \cite{edgcumbe_augmented_2016,kwartowitz2006toward}. This is largely due to inaccuracies in the reporting of the first five setup joints. To overcome this kinematics error, we propose a calibration method to find the API correction: $\boldsymbol{T}_{cor}$. This transform simultaneously accounts for errors in the hand-eye matrix, which itself depends on accurate API reporting.

To perform kinematics error correction,
the tip of the large needle driver instrument is touched to fiducials on the surgical training task, which, for simplicity, are the four corners of each ArUco marker. The 3D fiducial points, $\boldsymbol{p}^s$, are projected to 2D image coordinates using the camera-to-scene registration described in Section~\ref{subsec:scene_reg} such that $\boldsymbol{p}^{c,ac}={}^{c}\boldsymbol{T}_{s}\;\boldsymbol{p}^s$. These points are treated as the ground truth for calibration. The corresponding API reported tool-tip points are also projected onto the 2D image coordinates: $\boldsymbol{p}^{c,rep}={}^{ecm}\boldsymbol{T}_{lc}^{-1}\;{}^{ecm}\boldsymbol{T}_{psm}\;\boldsymbol{p}^{psm}$. The solution to $\boldsymbol{p}^{c,ac}=\boldsymbol{T}_{cor}\;\boldsymbol{p}^{c,rep}$ is then computed using our implementation of the Kabsch-Umeyama algorithm \cite{kabsch1976solution,umeyama1991least} combined with random sample consensus (RANSAC) for outlier rejection. In our experiments, we collected 8 sets of corresponding points to solve for $\boldsymbol{T}_{cor}$, balancing accuracy with calibration speed. Our API error correction algorithm can be found at: \href{https://github.com/AlexandreBanks6/dV-STEAR_Public.git}{dV-STEAR\_Public.git}.

\subsection{Camera and Hand-Eye Calibrations} \label{subsec:cam_calib}
The camera's intrinsics and 
lens distortion parameters are embedded in the scene registration and projection matrices. These parameters are estimated using a calibration method based on \cite{zhang2000flexible}. Forty images of an $8\times8$ checkerboard pattern were captured at varying distances and angles, and monocular calibration was performed using OpenCV in Python.

To compute the rigid hand-eye transformations, ${}^{ecm}\boldsymbol{T}_{lc}$ and ${}^{ecm}\boldsymbol{T}_{rc}$ shown in Fig. \ref{fig:playback}a., we created an open-source (\href{https://github.com/AlexandreBanks6/dV-STEAR_Public/blob/main/src/CameraCalibration_Automated_Best.py}{CameraCalibration.py}) calibration method for the dVRK based on the dual-quaternion approach proposed in \cite{malti_robust_2010}. During this calibration we use the same ArUco markers shown in Fig. \ref{fig:scene_reg}. 

We estimate the hand-eye transform for each camera, ${}^{ecm}\boldsymbol{T}_{c}$, using the classical hand-eye calibration equation: $\boldsymbol{A}{}^{ecm}\boldsymbol{T}_{c}={}^{ecm}\boldsymbol{T}_{c}\boldsymbol{B}$. Matrix $\boldsymbol{A}$ is composed by stacking $i=1..N-1$ camera motions, where each $\boldsymbol{A}_i={}^{c}\boldsymbol{T}_{s,i}\;{}^{c}\boldsymbol{T}_{s,i+1}^{-1}$ represents the change in camera pose with respect to the dry-lab surgical task. Similarly,  $\boldsymbol{B}$ is composed of ECM motions $\boldsymbol{B}_i={}^{base}\boldsymbol{T}_{ecm,i}^{-1}\;{}^{base}\boldsymbol{T}_{ecm,i+1}$. Each motion pair, $\boldsymbol{A}_i$ and $\boldsymbol{B}_i$, are $4 \times 4$ transforms that are vertically stacked into $((N-1) \times 4) \times 4$ matrices $\boldsymbol{A}$ and $\boldsymbol{B}$. ${}^{ecm}\boldsymbol{T}{c}$ is estimated using a dual quaternion method that finds rotation and translation simultaneously with RANSAC for robust outlier rejection.

\subsection{User Study Design}

An $24$-participant user study evaluated the impact of dV-STEAR compared to exploratory learning alone on novice performance in two tasks from the Fundamentals of Laparoscopic Surgery \cite{smith_fundamentals_2014}. The single-handed wire-chaser and pick-and-place tasks were selected as they train fundamental RAS skills and are indicative of operating room performance \cite{aghazadeh_performance_2016,jarc_face_2015}. Demonstrations were from an expert surgeon who has performed more than 100 procedures using the da Vinci Surgical System at Vancouver General Hospital. This expert practiced each task over 10 times and until performance plateaued before their motions were recorded. 

A between-subject design was used, with participants randomly assigned to the control group (No AR, $N=12$) or the experimental group (AR, $N=12$). A strict protocol was followed where participants first completed six familiarization sessions with the dVRK using a different task --- the peg transfer \cite{smith_fundamentals_2014}. Following familiarization, the following was repeated for both the wire-chaser and pick-and-place tasks: 1) five practice sessions with or without AR (depending on group), 2) a five-minute break, 3) five evaluation trials without AR during which performance was recorded, and 4) another five-minute break to complete the NASA Task Load Index (NASA-TLX) Questionnaire \cite{hart_nasa-task_2006}. During practice sessions, the experimental group shadowed AR demonstrations, while the No AR group learned through exploratory practice, as is typical in dry-lab surgical training.

\begin{figure}[h!]
\begin{center}    \includegraphics[width=\columnwidth,trim={1.95cm 14cm 17.9cm 0.195cm},clip]{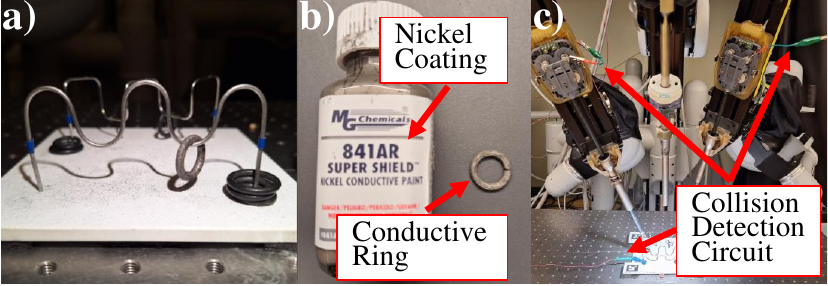}
    \caption{\textbf{a)} Single-handed wire-chaser task. Tape fixed to the right of the task indicates the start and end points. Tape fixed to the left marks the half-way point of the task; \textbf{b)} Nickel coating painted onto 3D printed ring; \textbf{c)} Cables of the collision detection circuit: two attached to the ground of the large needle driver instruments, one attached to the task's metal wire.}    \label{fig:task_images}
\end{center}
\end{figure}

The single-handed wire-chaser task (Fig. \ref{fig:task_images}a.) required moving a ring along a wire and back while minimizing collisions, training hand positioning and fine instrument control. Users were instructed to balance completion time with number of collisions. A piece of tape was fixed to the task (Fig. \ref{fig:task_images}a.) and recording started/ended when the ring visually passed this marker. Collisions were detected using an electrical circuit (Fig. \ref{fig:task_images}c.), which recorded when the nickel-coated 3D-printed ring (Fig. \ref{fig:task_images}d.) touched the wire. An Arduino Mega transmitted collision data over a serial interface to a PC which synchronized this with system time and instrument, ${}^{s}\boldsymbol{T}_{psm}$, pose. 

During the pick-and-place task (Fig. \ref{fig:scene_reg}), participants grasped small beads and place them on top of 10 posts in any order. Only one attempt per post was allowed, and attempts were successful only if the bead remained on the post. Participants could not retry a failed attempt, and were instructed to balance completion time with errors made. This task trained clutching and hand-eye coordination. The workspace included six posts in the front and four in the back. Data collection began when the participant moved the first bead above the top edge of the basket (Fig. \ref{fig:scene_reg}) and ended when the final bead was released.

Task order (wire-chaser vs. pick-and-place) was pseudo-randomized using a Latin Square design \cite{avinash2020evaluation}, ensuring each permutation of tasks occurred equally across the research population. AR and No AR groups received identical instructions, including images showing proper hand positioning and clutch use, as well as a video demonstration from the expert. The University of British Columbia Behavioural Research Ethics Board approved this study under \#H24-01947.

\section{Results}
\label{sec:Results}
dV-STEAR was analyzed for tool tracking accuracy and impact on novice performance, with Section \ref{subsec:results_pose_estim} quantifying the PSM pose estimation error, Section \ref{subsec:results_transformationchain_errors} detailing accuracy of the rendering pipeline, and Section \ref{subsec:results_userstudy} giving the results of the 24-participant comparison study. 

\subsection{Pose Estimation Accuracy}\label{subsec:results_pose_estim}
\begin{figure*}[!t]
\begin{center}
    \includegraphics[width=\textwidth,trim={1.25cm 0cm 1.7cm 0.8cm},clip]{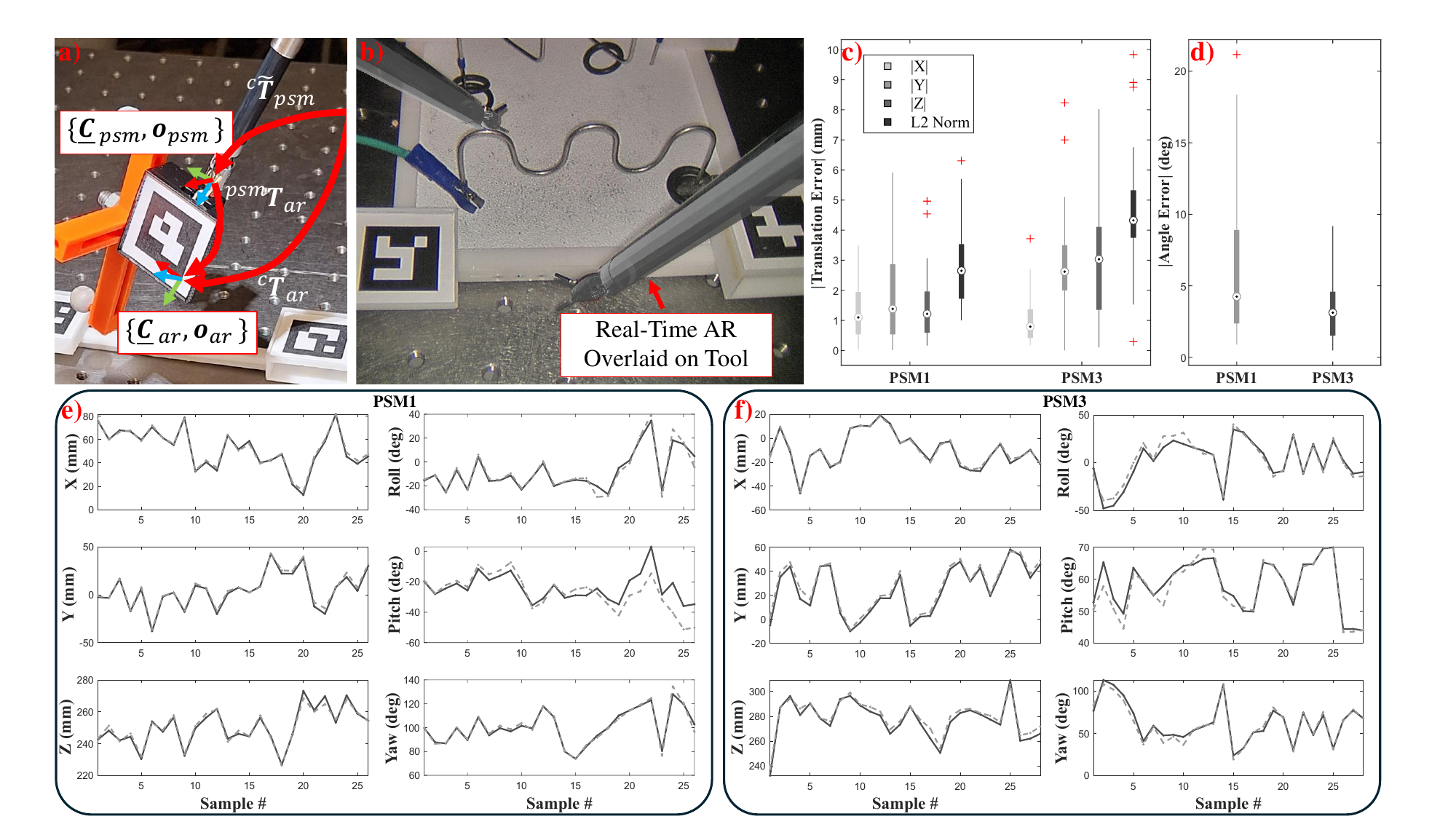}
    \caption{\textbf{a)} Transformations used to evaluate pose estimation accuracy. ${}^{c}\boldsymbol{T}_{ar}$ is the ground-truth, ${}^{c}\tilde{\boldsymbol{T}}_{psm}$ is the estimated PSM pose using our method, and ${}^{psm}\boldsymbol{T}_{ar}$ is the known rigid transform from the PSM to the ArUco marker; \textbf{b)} Example of real-time virtual overlay based on estimated pose; \textbf{c)} and \textbf{d)} Box plots of translational and angular errors in PSM pose; \textbf{e)} and \textbf{f)} Examples of position and orientation tracking using the pose estimation method (solid line) compared to the ground-truth (dotted line).}
    \label{fig:results_pose_estim}
\end{center}
\end{figure*}

To validate tool tracking accuracy during expert motion recordings, pose estimation errors were quantified over $N=26$ samples for each PSM1 and PSM3. Accuracy was computed by picking up an ArUco marker attached to the \textit{Stable Grasp} object developed in \cite{avinash_pickup_2019} and comparing visual tracking to our pose estimation pipeline. The baseline camera-to-ArUco transform, ${}^{c}\boldsymbol{T}_{ar}$ in Fig. \ref{fig:results_pose_estim}a., was found by extracting the four ArUco corners and solving the PnP problem using MAGSAC. The estimated ArUco pose, ${}^{c}\tilde{\boldsymbol{T}}_{ar}={}^{c}\tilde{\boldsymbol{T}}_{psm}\;{}^{psm}\boldsymbol{T}_{ar}$, was determined from our estimated PSM pose, ${}^{c}\tilde{\boldsymbol{T}}_{psm}$, and the known geometry of the ArUco object, ${}^{psm}\boldsymbol{T}_{ar}$. We computed ${}^{c}\tilde{\boldsymbol{T}}_{psm}$ by pre-multiplying Equation \eqbracketref{eqn:PoseRecording} with the registration transform ${}^{c,i}\boldsymbol{T}_{s}$. Here, the left camera was used as the reference.

The mean L2 error was $3.06\pm1.65$mm (PSM1) and $4.66\pm2.04$mm (PSM3). Per-axis errors are shown in Table \ref{table:pose_estim}, with an average absolute per-axis error of $1.55\pm1.24$mm (PSM1) and $2.32\pm1.79$mm (PSM3). Examples of position estimates for PSM1 and PSM3 are given in Fig.~\ref{fig:results_pose_estim}e. and Fig.~\ref{fig:results_pose_estim}f, illustrating that the translation signal closely follows the baseline visual tracking. Box plots of absolute translation error are given in Fig. \ref{fig:results_pose_estim}c. PSM1 errors across all axes were similar, whereas for PSM3, the x-axis error was lower than y- and z-axis errors. 

Orientation error was quantified by converting the rotational parts of ${}^{c}\tilde{\boldsymbol{T}}_{ar}$ and ${}^{c}\boldsymbol{T}_{ar}$ to quaternions and finding their angular difference. Angular error for PSM1 was $6.53\pm5.71$\textdegree and for PSM3 was $3.44\pm2.30$\textdegree. Examples of Euler angles for the baseline and estimated transforms are given in \ref{fig:results_pose_estim}f. 

A qualitative analysis was performed by overlaying the semi-transparent ghost tools over the tracked instruments in real-time. As shown in Fig. \ref{fig:results_pose_estim}b., the estimated pose of the ghost tools closely matched the actual instruments.

\begin{table}[h!]
\begin{center}
\caption{Pose Estimation Error}
\label{table:pose_estim}
\begin{tabular}{lll}
\hline\hline
      & \textbf{PSM1} & \textbf{PSM3} \\
\hline
\(|\)X\(|\) Error (mm) & 1.33 $\pm$ 0.93 & 1.04 $\pm$ 0.85\\
\(|\)Y\(|\) Error (mm) & 1.79 $\pm$ 1.57 & 2.97 $\pm$ 1.70\\
\(|\)Z\(|\) Error (mm) & 1.52 $\pm$ 1.24 & 2.96 $\pm$ 1.93\\
L2 Error (mm)& 3.06 $\pm$ 1.65 & 4.66 $\pm$ 2.04\\
Angle Error (\textdegree)& 6.53 $\pm$ 5.71 & 3.44 $\pm$ 2.30\\
\hline
\end{tabular}
\end{center}
\end{table}

\subsection{Transformation Pipeline Error Quantification}\label{subsec:results_transformationchain_errors}
Calibration parameters and other transformations in the recording and playback pipelines were analyzed for error contributions. The $40$-checkerboard camera calibration resulted in a re-projection error of $0.027\pm0.003$ pixels for the left camera and $0.032\pm0.004$ for the right. The hand-eye transformations, ${}^{ecm}\boldsymbol{T}_{lc}$ and ${}^{ecm}\boldsymbol{T}_{rc}$, were calibrated on 31 ECM motions using methods in Section~\ref{subsec:cam_calib} and evaluated on a separate set of 10 ECM motions with the resulting errors given in Table~\ref{table:camcalib_handeye}. The hand-eye transform error averaged across both cameras is $9.34\pm5.30$mm and $2.29\pm1.21$\textdegree. This reflects the raw hand-eye transform accuracy with embedded API reporting errors.

\begin{table}[h!]
\begin{center}
\caption{Camera Calibration and Hand-Eye Error}
\label{table:camcalib_handeye}
\begin{tabular}{lll}
\hline\hline
      & \textbf{Left Camera} & \textbf{Right Camera} \\
\hline
\textbf{Camera Calibration}\\
Reproj. Error (pix)& 0.027 $\pm$ 0.003 & 0.032 $\pm$ 0.004\\
\hline
\textbf{Hand-Eye}\\
L2 Error (mm)& 11.95 $\pm$ 5.58 & 6.78 $\pm$ 3.69\\
Angle Error (\textdegree)& 2.75 $\pm$ 1.37 & 1.83 $\pm$ 0.87\\
\hline
\end{tabular}
\end{center}
\end{table}

The robot kinematics error correction term, $\boldsymbol{T}_{cor}$, was computed for both cameras and PSM1/PSM3 instruments using methods in Section~\ref{subsec:error_corr}. Correction transforms were fit on 8 points and errors were computed with a different test dataset of 8 points (Table \ref{table:api_errorcorrection}). The average error in the API correction across both cameras and instruments was $0.55\pm0.62$mm.

\begin{table}[h!]
\begin{center}
\caption{dVRK API Correction Error}
\label{table:api_errorcorrection}
\begin{tabular}{lll}
\hline\hline
      & \textbf{Left Camera} & \textbf{Right Camera} \\
\hline

PSM1 L2 Error (mm)& 0.15 $\pm$ 0.08 & 0.21 $\pm$ 0.10\\
PSM3 L2 Error (mm)& 0.92 $\pm$ 0.71 & 0.92 $\pm$ 0.73\\
\hline
\end{tabular}
\end{center}
\end{table}

\begin{figure}[h!]
\begin{center}    \includegraphics[width=\columnwidth,trim={0.3cm 8.2cm 14.1cm 0.2cm},clip]{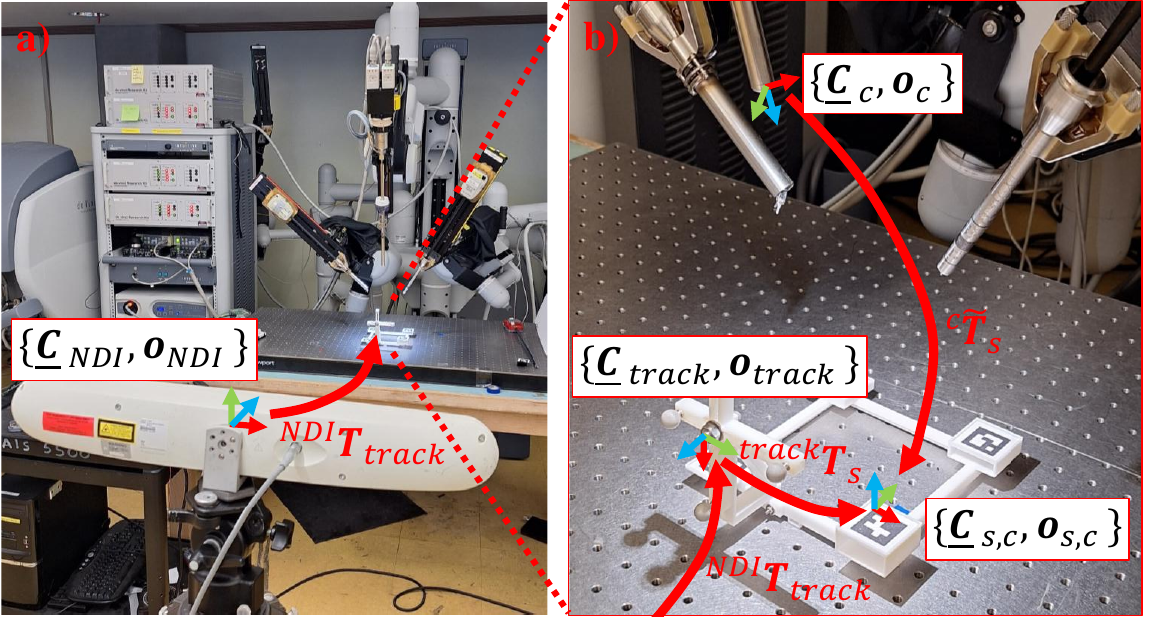}
    \caption{Transformations used to evaluate the scene registration transform ${}^{c}\tilde{\boldsymbol{T}}_{s}$. ${}^{NDI}\boldsymbol{T}_{track}$ is the pose estimated from the NDI tracker. ${}^{track}\boldsymbol{T}_{s}$ represents the known rigid geometry converting the tracker frame, $\{\underline{\boldsymbol{C}}_{track},\boldsymbol{o}_{track}\}$, to the scene's coordinate frame, $\{\underline{\boldsymbol{C}}_{s},\boldsymbol{o}_{s}\}$.}    \label{fig:scene_registration_validation_transforms}
\end{center}
\end{figure}

Scene registration transforms, ${}^{lc,i}\boldsymbol{T}_{s}$ and ${}^{rc,i}\boldsymbol{T}_{s}$, from Section~\ref{subsec:scene_reg} were validated for relative accuracy against an NDI tracker (Fig. \ref{fig:scene_registration_validation_transforms}.) attached to the registration target. The NDI tracker frame, $\{\underline{\boldsymbol{C}}_{track},\boldsymbol{o}_{track}\}$, was moved through $N=29$ orientations and positions, with the initial location recorded as the base pose. The registration transform, ${}^{c}\tilde{\boldsymbol{T}}_{s}$, motion relative to the base pose was compared to the NDI transform, ${}^{NDI}\boldsymbol{T}_{s}={}^{NDI}\boldsymbol{T}_{track}\;{}^{track}\boldsymbol{T}_{s}$, motion relative to the base pose. The transform from the NDI tracker to the scene, ${}^{track}\boldsymbol{T}_{s}$, was known from the CAD model geometry. Scene registration errors are provided in Table \ref{table:scene_registration}, where an average L2 error of $8.25\pm4.39$mm and an average angle error of $5.12\pm3.16$\textdegree were found across both left and right cameras.

\begin{table}[h!]
\begin{center}
\caption{Scene Registration Error}
\label{table:scene_registration}
\begin{tabular}{lll}
\hline\hline
      & \textbf{Left Camera} & \textbf{Right Camera} \\
\hline
\(|\)X\(|\) Error (mm) & 4.34 $\pm$ 3.51 & 4.39 $\pm$ 3.62\\
\(|\)Y\(|\) Error (mm) & 0.99 $\pm$ 0.95 & 1.07 $\pm$ 1.05\\
\(|\)Z\(|\) Error (mm) & 3.73 $\pm$ 3.08 & 7.96 $\pm$ 4.65\\
L2 Error (mm)& 6.56 $\pm$ 3.61 & 9.94 $\pm$ 4.50\\
Angle Error (\textdegree)& 2.84 $\pm$ 1.80 & 7.40 $\pm$ 2.53\\
\hline
\end{tabular}
\end{center}
\end{table}

\subsection{User Study}\label{subsec:results_userstudy}

The $24$ novices enrolled in the study had a mean age of $26.2\pm4.3$ years, and only had $2.6\pm5.6$ prior uses with a surgical robot. $11$ participants self-identified as female, $12$ as male, and $1$ as non-binary. $13$ self-identified as having a European cultural background, $6$ with East Asian, and one with each of Hispanic, Middle Eastern, Southeast Asian, South Asian, and African background. $18$ participants were right handed, $5$ left handed, and one ambidextrous. 

Statistical analysis was performed with MATLAB R2023b and a two-stage approach to calculate significance. In this method, data was first checked for normality using the Shapiro-Wilk test at a $0.05$ significance level. If not normally distributed, the null hypothesis was tested using the Wilcoxon rank sum test. For normally distributed data, comparison groups were then assessed for homogeneity of variance. If variances were equal, the null hypothesis was tested using the independent t-test; otherwise Welch's method was used.

\subsubsection{Single-Handed Wire-Chaser Task}

\begin{figure}[h!]
\begin{center}
    \includegraphics[width=\columnwidth,trim={1.41cm 10.7cm 5cm 1cm},clip]{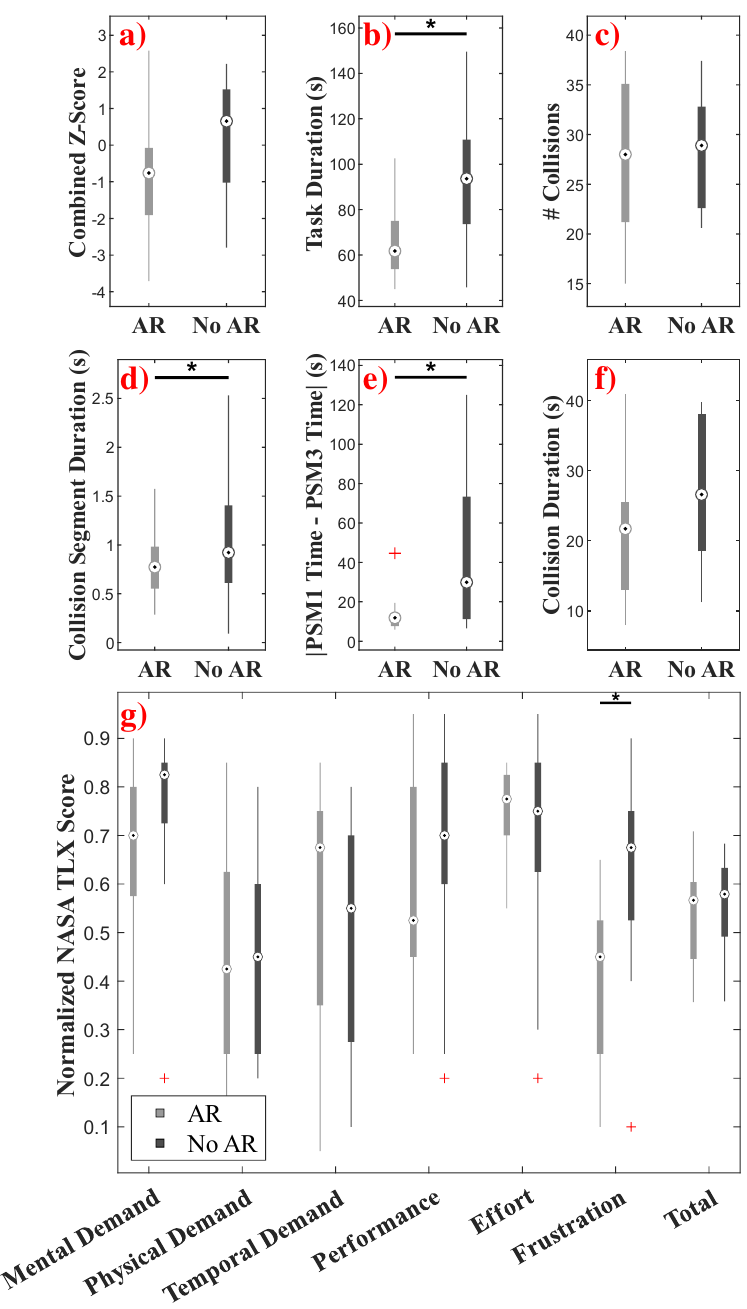}
    \caption{User study ($N=24$) results comparing AR and No AR groups on the single-handed wire-chaser task. \textbf{a)} Combined z-score computed by summing standardized task duration, number of collisions, and collision duration; \textbf{b)} and \textbf{c)} Task duration and number of collisions; \textbf{d)} Average duration of each collision; \textbf{e)} Absolute difference in amount of time that PSM1 and PSM3 were used; \textbf{f)} Total time that wire touched the ring over a single trial; \textbf{g)} Qualitative NASA-TLX results.}
    \label{fig:userstudy_onehanded_results}
\end{center}
\end{figure}

During the wire-chaser task, participants were evaluated on several quantitative metrics. For task duration (Fig. \ref{fig:userstudy_onehanded_results}b.), the AR group completed the task significantly faster ($p=0.03$) with a completion time of $65.73\pm16.97$s compared to $98.50 \pm43.09$s for the No AR group. The number of collisions were similar between groups ($p=0.82$), with AR participants having $27.63 \pm8.13$ collisions per trial compared to $30.33\pm11.02$ for the No AR group. However, average duration of collision segments --- the average time users dragged the ring along the wire per collision --- was significantly less ($p=0.01$) for the experimental group ($0.80\pm0.33$s) compared to controls ($1.08\pm0.62$s) as illustrated in Fig. \ref{fig:userstudy_onehanded_results}d.

The \textit{hand balance error} metric, defined as the absolute difference in time using PSM1 and PSM3, was much lower for the AR group than the No AR group (AR=$13.91\pm13.89$, No AR=$44.51\pm46.09$, $p=0.04$). Additionally, collision duration --- the total time the wire touched the ring --- was lower for the experimental group ($20.96\pm9.14$) compared to controls ($30.74\pm17.69$). The AR group performed better in the primary outcome: a composite z-score \cite{cheadle_analysis_2003} computed by summing the standardized task duration, number of collisions, and collision duration. However, z-score differences were not significant (AR: $-0.93\pm1.74$; No AR: $0.93\pm3.29$; $p=0.08$). A lower z-score indicates better overall performance.

NASA-TLX results, Fig. \ref{fig:userstudy_onehanded_results}g., give qualitative metrics on self-reported workload. The AR group reported significantly lower ($p=0.01$) frustration levels than their No AR counterparts. Although not significant, the AR group reported lower mental demand, physical demand, and total workload. Self-rated performance was lower in the AR group ($p=0.58$).

\subsubsection{Pick-and-Place Task}

\begin{figure*}[!t]
\begin{center}
    \includegraphics[width=\textwidth,trim={1.6cm 2.1cm 2.85cm 4.2cm},clip]{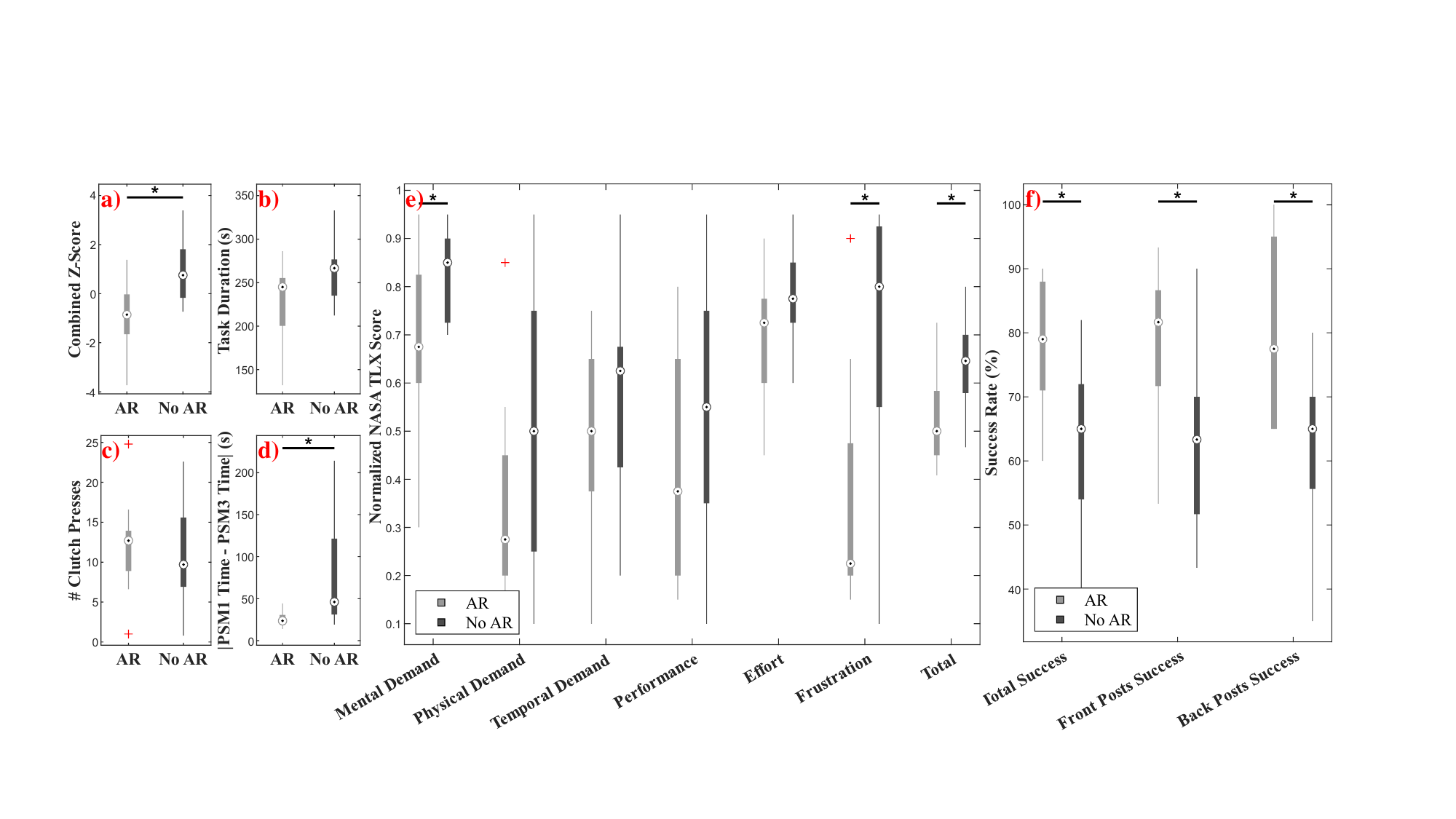}
    \caption{User study (N=24) results comparing AR and No AR groups on the pick-and-place task. \textbf{a)} Combined z-score computed by summing standardized task duration and total success rate; \textbf{b)} and \textbf{c)} Task duration and number of clutch presses; \textbf{d)} Absolute difference in time PSM1 and PSM3 were used; \textbf{e)} NASA-TLX results; \textbf{f)} Success rate defined as number of beads resting on posts at end of trial. Decomposed into $6$ front posts, $4$ back posts, and all $10$ posts. Users had one attempt per post.}
    \label{fig:userstudy_pickandplace_results}
\end{center}
\end{figure*}

Fig. \ref{fig:userstudy_pickandplace_results} shows both the quantitative and qualitative results from the pick-and-place task. The AR group achieved a total success rate of $78.5\pm10.0$\% which was significantly higher ($p=0.004$) than the $63.0\pm13.1$\% success rate of the No AR group. When decomposing the success rates, the experimental group performed significantly better for the back posts ($p=0.02$) and front posts ($p=0.02$).

The primary outcome was the composite z-score, which was computed by combining standardized success rate and task duration. Participants in the AR group achieved a significantly lower z-score ($-0.923\pm1.53$,$p=0.005$) than the control group (z-score=$0.963\pm1.40$). Additionally, \textit{hand-balance} was better ($p=0.004$) in the AR group ($25.78\pm9.35$s) compared to the No AR group ($75.36\pm63.52$s). 

Although not statistically significant ($p=0.07$), participants using AR completed the task faster ($228.65\pm47.06$s) than controls ($263.22\pm37.75$s). Number of clutch presses were similar across groups, with AR participants averaging $11.97\pm5.81$ compared to $11.22\pm6.57$ for the No AR group.

NASA-TLX results show that the AR group self-reported feeling significantly less mental demand ($p=0.03$) and frustration ($p=0.007$) than their counterparts. Total workload was also rated significantly lower ($p=0.004$) for the experimental group. Moreover, AR participants rated their experience lower for physical demand ($p=0.16$), temporal demand ($p=0.26$), performance ($p=0.23$), and effort ($p=0.10$).

\section{Discussion}
\label{sec:Discussion}

AR has been shown to help 3D understanding of gross anatomy \cite{bork_effectiveness_2021} and in conveying spatial relationships \cite{rossano_augmented_2020}. In addition, AR impacts learning and retention in many settings, such as education \cite{alkhabra2023augmented} and assembly line environments \cite{gavish_evaluating_2015}. Given its benefits and widespread use in other fields, the lack of AR integration in robotic surgery training underscores the need for further technological advancements and comprehensive evaluation of these systems. This paper proposes and validates an AR system that enables experts to record instrument motions and have them played back to novices as demonstrations in any robot configuration. 

While previous research has explored AR-based telementoring where ghost tools directly guide trainees \cite{jarc_proctors_2017,matu_stereoscopic_2014,jarc_beyond_2016,shabir_preliminary_2022,casas-yrurzum_new_2023,shabir_towards_2021}, these systems do not have functionality for asynchronous expert demonstration. Other studies have recorded expert hand motions during open surgery \cite{cota_neto_what_2024} and captured video and tactile data during RAS for AR playback \cite{krauthausen_robotic_2021}. Though shown to be qualitatively beneficial, these methods depend on a fixed robot setup where motions are recorded and played back in a static reference frame. Reinforcement learning (RL) has also been used to provide AR guidance during RAS training \cite{long_integrating_2022}. While novices were able to closely follow the RL guidance in this study with sub-pixel accuracy, this system has several drawbacks, including that the training task must be placed in the same position with respect to the robot base, ECM motion was not accounted for, and only 2D guidance trajectories were generated. In contrast, our work introduces a kinematic-based AR system that records full 6 DoF instrument movements aligned to the surgical training task and replays them to novices irrespective of the robot's setup joint positions. 

Through robust error quantification, we demonstrated that our AR system accurately tracks surgeon motions. The pose estimation translation error was $3.06\pm1.65$mm for PSM1 and $4.66\pm2.04$mm for PSM3, yielding a combined error of $3.86\pm2.01$mm. Additionally, our system achieved an average angular error of $4.99\pm4.58$\textdegree. For comparison, vision-based pose tracking methods report similar errors. A particle filter approach \cite{hao2018vision} achieved a position error of $2.6$mm and an orientation error of $3.8$\textdegree. Another study \cite{jiang2023development}, showed that keypoint detection-based pose estimation had a positional error of $1.4$mm and an orientation error of $2.9$\textdegree. Given that our markerless kinematics-based tracking method performs within $1-2.5$mm and $2$ degrees of these state-of-the-art image-based methods, and considering that mentees are able to follow AR during telementoring with a maximum positional accuracy of $3.39\pm0.76$mm \cite{shabir_preliminary_2022}, we conclude that our tracking error falls within acceptable bounds for AR-RAST.

Our user study (N=24) shows that AR playback led to both quantitative and qualitative improvements for novices during RAS training. During the single-handed wire-chaser task, novices using dV-STEAR completed the task faster ($p=0.03$), with a time of $65.73\pm16.97$s compared to $98.50\pm43.09$s for the control group. Given that the expert surgeon completed the task in $46.9$ seconds, this suggests that AR playback helps novices to better match expert pacing. Importantly, the incrased speed of the experimental group did not lead to more collisions as shown in Fig. \ref{fig:userstudy_onehanded_results}c. Additionally, the average time of contact during collisions was smaller for the AR group by $0.47\pm0.50, p=0.01$s, indicating improved spatial awareness, which is a known benefit of AR teaching tools \cite{bork_effectiveness_2021}. Novices using AR also demonstrated significantly better use of both hands, which suggests that seeing an expert use both tools promotes improved hand balance in trainees. Although not statistically significant, the AR group performed better across all other quantitative metrics, including the combined z-score and total collision time, further highlighting the benefits of our record-and-playback approach for RAS training.

Despite indicating a similar total workload on the NASA-TLX survey, the AR group self-reported feeling significantly less frustration during the wire-chaser task compared to controls. This corroborates with prior research \cite{gomez2023analysis} showing that AR can reduce negative emotions such as frustration and anxiety in educational environments. 

For the pick-and-place task, AR participants achieved a significantly higher success rate by $15.54\pm21.46$\% ($p=0.004$). The composite z-score for the AR group was also significantly better ($p=0.005$). These improvements may be due to the AR group adopting a similar instrument positioning strategy as the expert, which was observed to cause fewer occlusions of the bead by the tool tip. Similar to the wire-chaser task, the AR group exhibited better hand balance (AR=$25.78\pm9.35$s, No AR=$75.36\pm63.52$s, p=0.004). This aligns with findings from previous studies \cite{feng_virtual_2018} showing that AR improves movement economy, particularly for the non-dominant hand.

NASA-TLX results for the pick-and-place task indicate the AR group found the task significantly less frustrating and mentally demanding and had a significantly lower total workload. Reduction in mental demand aligns with previous findings that AR-based learning environments lower mental strain \cite{bork_effectiveness_2021}. Interestingly, self-rated performance was lower in the AR group for both the pick-and-place and wire-chaser tasks. This could be due to the AR group using the expert surgeon's proficiency as their comparison baseline. Such a perception may be beneficial as it could lead to higher motivation in trainees, a known effect of AR systems in teaching \cite{gomez2023analysis}.

The quantitative and qualitative benefits of our novel record-and-playback system reflect trends observed in other AR-RAST systems. Work by Jarc et al. \cite{jarc_beyond_2016} found that both novices and experts preferred 3D instument overlays to other guidance methods for their helpfulness in surgical training. Compared to prior research on telementoring-based AR, our system is the first to provide a setup-invariant method to playback expert guidance. To the best of our knowledge, this is also the first work in RAS to robustly evaluate an AR system's tracking accuracy and perform quantitative and qualitative analysis of performance improvements through a user study, marking a key advancement in the AR-RAST field.

As a follow-up to our work, a longitudinal study could assess the impact of dV-STEAR on long-term skill improvement as prior research \cite{alkhabra2023augmented} has shown that AR enhances learning retention. Additionally, future studies could enroll surgical residents and evaluate the impact of AR training on their operative performance through measures such as the Global Evaluative Assessment of Robotic Skills \cite{younes2023clinically}. Another important extension of our work would be to combine guidance signals from different expert surgeons, potentially through methods such as learning from demonstration \cite{su2021toward}. Furthermore, our user study design could be expanded by including another experimental group that received AR guidance during both the training and testing phases. This would shed light on whether continuous AR exposure provides benefits compared to training-only exposure.

From a technical perspective, future iterations of dV-STEAR could incorporate algorithmic, rendering, and measurement improvements to further enhance the AR experience. Changes in the remote center of motion (RCM) position across setups may cause misalignment between the shaft of the AR rendering and that of the novice's instrument. A potential solution could involve using machine learning to dynamically adjust the overlay's RCM position while preserving the 6 DoF pose of the playback. Additionally, the AR experience may be enhanced by coupling it with haptic feedback --- an approach previously implemented on the dVRK for ultrasound probe guidance \cite{moore2024enabling}. Furthermore, individual differences in eye disparity affect depth perception and visual fusion \cite{avinash2020evaluation}, suggesting a calibration method could be implemented to tailor AR depth to each user's eyes. We also believe that errors in PSM pose estimation shown in \ref{fig:results_pose_estim} are partially a due to measurement errors from inaccuracies in the rigid transform of the pick-up AruCo marker. These errors could be minimized by mounting a marker directly on the final joint of the instrument. Another possible improvement would be to increase the number of tasks that used with dV-STEAR to match VR platforms such as SimNow (Intuitive Surgical, Inc) which has 45+ exercises. Lastly, incorporating other AR objects alongside ghost instruments, such as needles and sutures, may enhance training and evaluation as shown in \cite{lahanas_novel_2015}.

Our study shows that setup-invariant AR exploiting teaching by demonstration improves task performance, enhances bi-manual hand balance, and reduces frustration and mental demand in novices. dV-STEAR aligns with future directions in the RAS education field, including reducing reliance on expert supervision \cite{sinha2023current}, improving AR alignment accuracy \cite{seetohul_augmented_2023}, and integrating AR systems into training curricula \cite{fu2023recent}. By providing expert-level instruction and visual guidance in a structured, self-directed environment, our system addresses a critical barrier in RAS training: the lack of infrastructure enabling novices to practice surgical skills outside the OR while benefiting from AR mentorship. This work paves the road for future studies to combine expert-independent AR training with other functionalities such as multi-modal playback and automatic performance evaluation.

\section{Conclusion}
\label{sec:Conclusion}
By enabling novices to develop skills outside the OR, practice through real-world exploration, and receive guidance without needing direct supervision, dV-STEAR overcomes issues in the portability of current AR-RAST systems. With a PSM pose estimation error of $3.86\pm2.01$mm, our system enables reliable expert motion recordings invariant to changes in the ECM position or the setup joints of the robot. During playback, we show that this form of AR guidance improves completion speed ($p=0.03$), duration of collisions ($p=0.01$), and hand balance in a path following task ($p=0.04$). In a precision pick-and-place task, we found that AR playback led to improved overall performance ($p=0.005$) and better hand balance ($p=0.004$). Beyond quantitative improvements, users receiving AR training indicated feeling less frustration and mental demand. A few important questions remain regarding dV-STEAR's impact on learning retention, intraoperative applicability, and the potential benefits of multi-modal information playback. Altogether, our study illustrates the technical viability of an open-source, setup-invariant, AR-RAST system (\href{https://github.com/AlexandreBanks6/dV-STEAR\_Public.git}{dV-STEAR\_Public.git}) and its impact on enhancing surgical skill performance in novices. This work tackles broader challenges facing surgical education and advances AR integration into RAS training.

\section*{Acknowledgments}
The authors gratefully acknowledge support from the NSERC Canada Graduate Scholarships and the C.A. Laszlo Biomedical Engineering Chair held by Professor Salcudean, as well as use of the da Vinci Research Kit with support from Intuitive Surgical and Johns Hopkins University. The authors have no conflicts of interest to disclose.

\bibliographystyle{IEEEtran}
\bibliography{dvFlexAR_Citations}

 
\vspace{11pt}

\vspace{-33pt}
\begin{IEEEbiography}[{\includegraphics[width=1in,height=1.25in,clip,keepaspectratio]{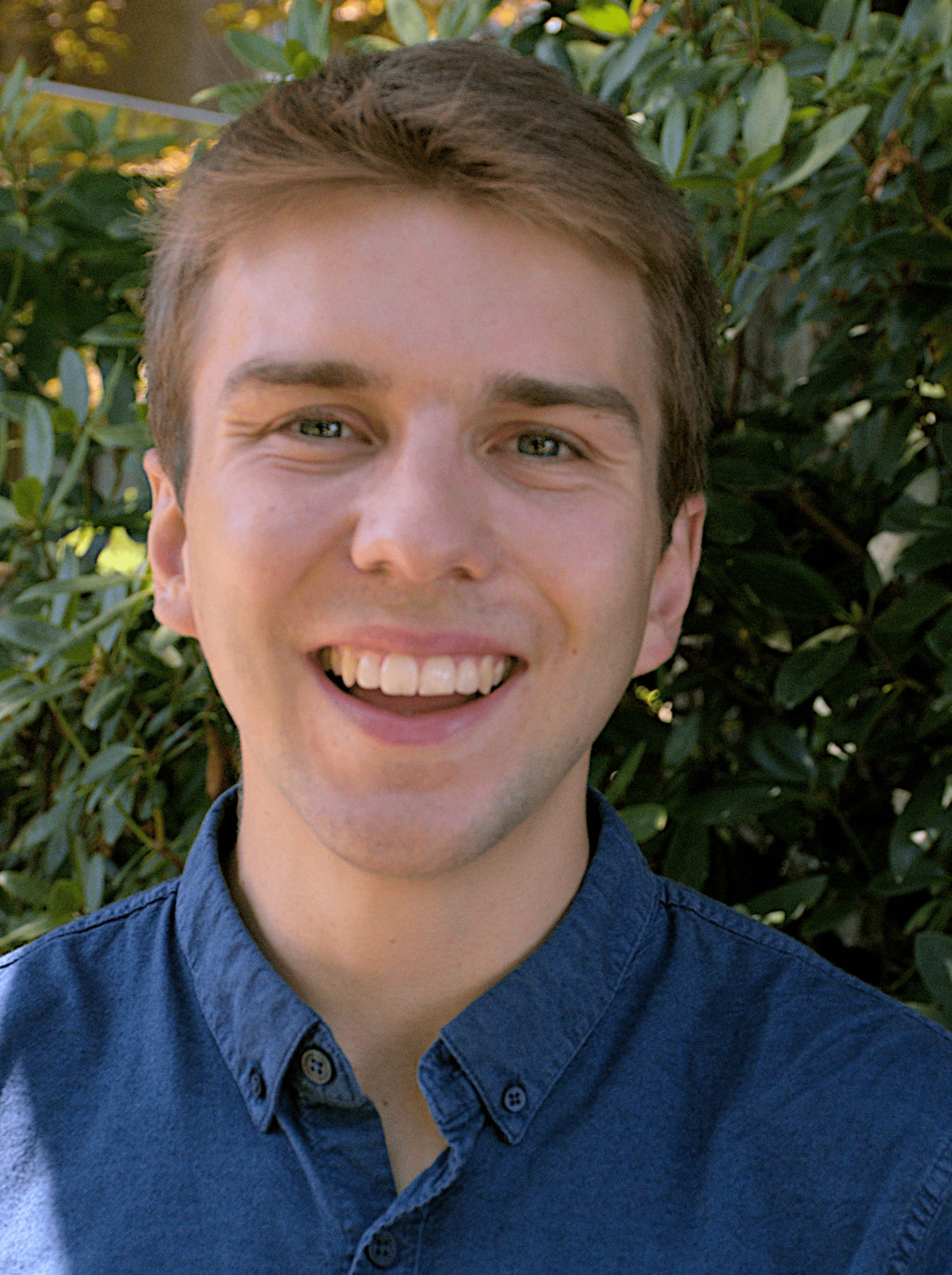}}]{Alexandre Banks} was born in Saint John, NB, Canada, in 1999. After graduating with a B.Sc. in Electrical and Computer Engineering from the University of New Brunswick (UNB), he moved to the University of British Columbia (UBC) where he is pursuing an M.A.Sc. in Biomedical Engineering. Alexandre completed research internships in rehabilitation robotics at the UNB Institute of Biomedical Engineering in 2020 and 2021, and from 2022-2025 worked in surgical robotics at the UBC Robotics and Control Laboratory. In 2025, Alexandre completed an internship at the University of Oxford on guidance and computer vision for fetal ultrasound.
\end{IEEEbiography}

\vspace{-33pt}
\begin{IEEEbiography}
[{\includegraphics[width=1in,height=1.25in,clip,keepaspectratio]{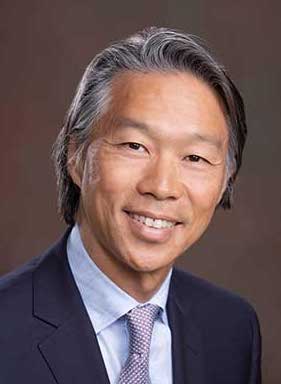}}]{Richard Cook}
After completing his cardiac surgery training at UBC, he completed a mitral valve repair fellowship with Professor Christophe Acar in Paris, followed by a robotic and minimally-invasive surgery fellowship with Dr. W. Randolph Chitwood, Jr., in North Carolina.
Dr. Cook is a Clinical Professor of Surgery at UBC and the Director of Minimal Access and Robotically-Assisted Cardiac Surgery at the Vancouver General Hospital. His primary research interests include minimally-invasive cardiac surgery, including mini-thoracotomy mitral valve repair and replacement, tricuspid valve repair and replacement, and atrial septal defect repair. He also interested in robotically assisted coronary artery bypass surgery and transcatheter management of valvular heart disease.

\end{IEEEbiography}

\vspace{-33pt}
\begin{IEEEbiography}
[{\includegraphics[width=1in,height=1.25in,clip,keepaspectratio]{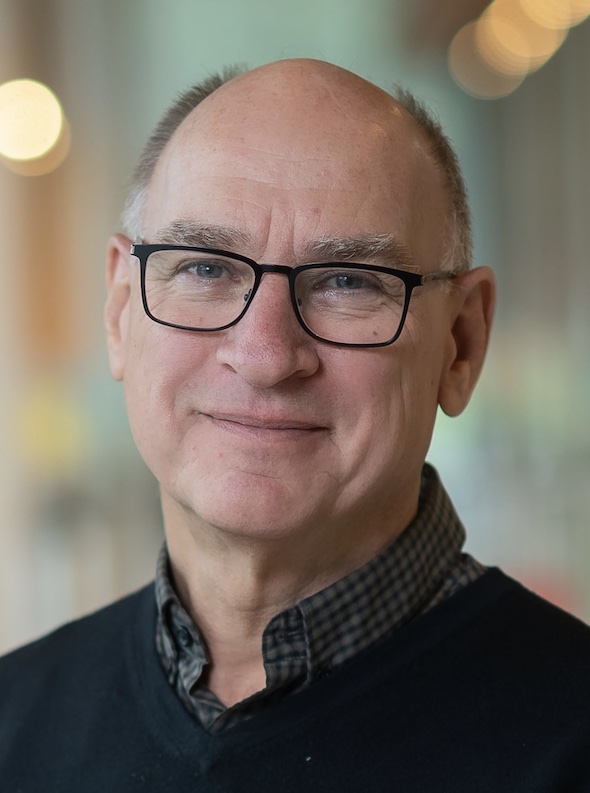}}]{Septimiu E. Salcudean} (Fellow, IEEE) was born in Cluj, Romania. He received the B.Eng. (Hons.) and M.Eng. degrees in electrical engineering from McGill University, Montreal, QC, Canada, in 1979 and 1981, respectively, and the Ph.D. degree in electrical engineering from the University of California at Berkeley, Berkeley, USA in 1986. He was a Research Staff Member with the IBM T. J. Watson Research Center from 1986 to 1989. He then joined UBC where he is currently a Professor with the Department of Electrical and Computer Engineering and the School of Biomedical Engineering. He holds the C.A. Laszlo Chair of Biomedical Engineering. He has been a Co-Organizer of the Haptics Symposium, a Technical Editor and Senior Editor of the IEEE Transactions on Robotics and Automation, and on the program committee of the ICRA, MICCAI, and IPCAI Conferences. He is currently on the Steering Committee of the IPCAI Conference and on the editorial board of The International Journal of Robotics Research. He is a Fellow of MICCAI and of the Canadian Academy of Engineering.

\end{IEEEbiography}

\vfill

\end{document}